# A visualization method for data domain changes in CNN networks and the optimization method for selecting thresholds in classification tasks

Minzhe Huang[1*]　　　　　　　Changwei Nie[2]　　　　　　　Weihong Zhong[3]
Akuvox　　　　　　　　　　　Akuvox　　　　　　　　　　　Akuvox
`minzhe.haung@akuvox.com`　`15256089750@163.com`　`weihong_2000@163.com`

## Abstract

*In recent years, Face Anti-Spoofing (FAS) has played a crucial role in preserving the security of face recognition technology. With the rise of counterfeit face generation techniques, the challenge posed by digitally edited faces to face anti-spoofing is escalating. Existing FAS technologies primarily focus on intercepting physically forged faces and lack a robust solution for cross-domain FAS challenges. Moreover, determining an appropriate threshold to achieve optimal deployment results remains an issue for intra-domain FAS. To address these issues, we propose a visualization method that intuitively reflects the training outcomes of models by visualizing the prediction results on datasets. Additionally, we demonstrate that employing data augmentation techniques, such as downsampling and Gaussian blur, can effectively enhance performance on cross-domain tasks. Building upon our data visualization approach, we also introduce a methodology for setting threshold values based on the distribution of the training dataset. Ultimately, our methods secured us second place in both the Unified Physical-Digital Face Attack Detection competition and the Snapshot Spectral Imaging Face Anti-spoofing contest. The training code is available at https://github.com/SeaRecluse/CVPRW2024.*

## 1. Introduction

Face recognition technology is currently widely used in areas such as access control, facial payment systems, and device unlocking. These applications all require validation of the face's reliability, namely determining whether it represents real face data. Traditional biometric identification attacks have mostly originated from real-world physical attacks, such as using 2D/3D printed objects for spoofing or high-definition electronic replay attacks. To ensure the reliable use of face recognition systems, a series of face anti-spoofing competitions[1][2][3] have been organized to enhance performance. There are also many effective FAS methods that have been proposed[4][5][6][7][8] . However, in recent years, with the rise of technologies like GAN[9] or Stable Diffusion[10], and the widespread use of applications like deepfake[11], digital editing of faces has posed a significant challenge to the authenticity of faces in images or videos. Consequently, a series of algorithms[12][13] designed to determine whether a face has been digitally tampered with have emerged. Considering both aspects as extensions of the FAS task, CVPRW2024 has created a dataset and established three tracks to study hybrid and cross-domain data, aiming to integrate interception techniques for physically forged and digitally edited faces.

To study the distribution of data across different domains and to measure the effectiveness of model training, we proposed a method for visual analysis using prediction results. Through validation, we discovered how techniques like downsampling and Gaussian blur enhance the model's generalization capabilities. Downsampling tends to enlarge the data domain, while Gaussian blur makes classes more cohesive within themselves. Additionally, using this visualization method, we found that traditional threshold determination methods based on metrics like ACER[14] might not be well-suited for deployment scenarios. Consequently, we also introduced an approach for determining thresholds based on our visualization analysis scheme.

## 2. Related Work

### 2.1. Data visualization tasks

Visualization of data is often performed to interpret and analyze the effects of a model. Existing visualization methods are mostly designed by depicting the areas of response on the model with respect to the data, such as with tools like Captum[15]. However, these methods do not provide direct guidance for optimizing Face Anti-Spoofing (FAS) tasks, nor do they reveal how out-of-domain data performs on the model.

### 2.2. Analysis of the role of data augmentation

The analysis of the effectiveness of data augmentation in deep learning tasks has been ongoing[16][17]. Although validations, such as downsampling and Gaussian blur, have proven to effectively enhance generalization, they have not been clearly demonstrated through quantitative analysis. Therefore, exactly how data augmentation operations specifically affect model performance remains a worthy subject for research.

---
[*]Corresponding author

### 2.3. Threshold selection in FAS tasks

In the FAS (Face Anti-Spoofing) domain, common methods for determining thresholds are usually based on the following performance metrics:

**ACER** (Average Classification Error Rate): The threshold when calculating equal error rates.

**ROC**[18] (Receiver Operating Characteristic): Calculation of a set of TPR-FPR pairs, using the threshold that meets a specific criterion when FPR-TPR is greater than a certain standard, for example using *FPR = 0.1%* and *TPR > 99%*.

The primary issue with the above threshold selection methods is that they calculate only proportionate values. In AI tasks, after training, the scores of difficult samples often differ greatly from those of easy ones. The identification of challenging samples typically relates to the strength of the model's generalization abilities. This is particularly critical for cross-domain tasks, where easy samples may not play a significant role.

## 3. Methodology

In this section, we will first define some basic concepts such as prediction center $Pred_{center}$, data radius $Radius$, and data density $Density$. These will assist in establishing our data visualization. Sec. 3.1 will introduce these concepts in detail and demonstrate the effects of data visualization. Sec. 3.2 will show the changes in the data domain after undergoing downsampling and Gaussian blur processes, and, in conjunction with the data distribution of Unified Physical-Digital Face Attack Detection, present our hypothesis. Sec. 3.3 will exhibit the effects under different threshold settings based on our visualized patterns and propose our method for setting thresholds. Lastly, we will validate our hypothesis through experiments in Sec. 4.

### 3.1. Select prediction center point

During the training of supervised tasks, we typically assign different label values for different classes. However, due to the calculation method of loss, it is usually the case that the loss cannot reach zero. Therefore, the label values do not represent the actual prediction center after model training. We need to determine the true prediction center based on the predicted scores from the training or validation set.

For a set of prediction scores, $Preds$, after we sort them from smallest to largest, there are several methods to determine the center value of $Preds$.

**Use mean value.** Eq. (1) This is probably the most common way.

$$Pred_{mean} = \frac{\sum_{n=1}^{N} Preds_n}{N} \quad (1)$$

**Use median values.** Eq. (2) This ensures that the amount of data on both sides of the prediction score is balanced.

$$Pred_{median} = \frac{Preds_{\lfloor \frac{N}{2} \rfloor} + Preds_{\lfloor \frac{N+1}{2} \rfloor}}{2} \quad (2)$$

**Use distance equal values.** Eq. (3) The sum of the distances between the two segments of the predicted score to reach this value is balanced. This takes into account difficult samples and some singular values.

$$Pred_{balance} = Pred_m$$

$$\sum_{n=1}^{m-1} Pred_m - Preds_n = \sum_{n=m+1}^{N} Preds_n - Pred_m \quad (3)$$

**Use weighted distance equal values.** Eq. (4) Similar to the selection of distance equal values, on top of this, the prediction score is additionally used as a weight for weighting. Doing so will amplify the impact of difficult samples far from the center point.

$$Pred_{weighted\ balance} = Pred_m$$

$$\sum_{n=1}^{m-1} (Pred_m - Preds_n)^2 = \sum_{n=m+1}^{N} (Preds_n - Pred_m)^2 \quad (4)$$

Assuming that we use any of the above calculation methods to obtain the $Pred_{center}$, use $Radius$ Eq. (5) to determine the radius of the prediction data domain, in the common FAS two-classification task, we can draw a two-dimensional prediction area on the predicted one-dimensional 0-1 coordinate axis.

$$Radius = Max(Pred_{center} - Min(Preds),$$
$$Max(Preds) - Pred_{center}) \quad (5)$$

Using randomly generated data samples of length 500 that conform to a Gaussian distribution with a standard deviation of 0.1, we have plotted the prediction region diagrams Figure 1 using the above four methods as $Pred_{center}$ points.

Taking into account computational complexity as well as the importance of focusing on difficult samples, in this paper we consistently use distance equal values as central points for visualization drawing.

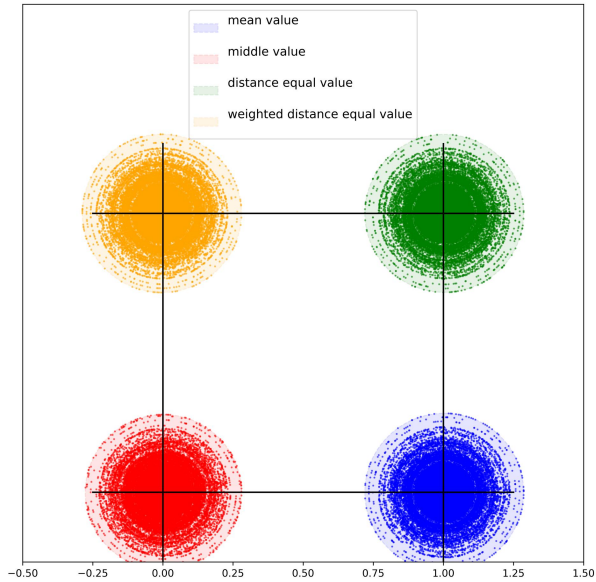

Figure 1: The differences in visualized images drawn using four types of confirmed Pred$_{center}$ are used. It can be observed that when taking into account the 'distance equal', the predicted distribution exhibits more pronounced stratified variability.

### 3.2. Study on Data Domain Variations

We conducted a study on how downsampling and Gaussian blur improve the model's generalization capabilities in Figure 2.

**Downsampling** reduces dimensionality and thus loses high-frequency information such as edges and textures, but it still preserves the core features; moreover, the noise introduced by downsampling can extend the boundaries of the original data domain.

**Gaussian blur**, by averaging the values of neighboring pixels, can help suppress noise and highlight underlying features, thereby making the original data domain more cohesive.

In the Unified Physical-Digital Face Attack Detection competition, we can observe that for FAS tasks, the Phys data domain and the Adv/Digital data domain are markedly different. If represented on a one-dimensional data axis ranging from 0 to 1, with 0 as the label for a real face, the data domains would be distributed from left to right as follows: Live - Adv/Digital - Phys. Therefore, if one wishes to generalize to other datasets using only one specific data domain, it is necessary to employ data augmentation techniques to diffuse the data domain boundaries. We then validated that downsampling and Gaussian blur as forms of data augmentation serve different purposes in Figure 3.

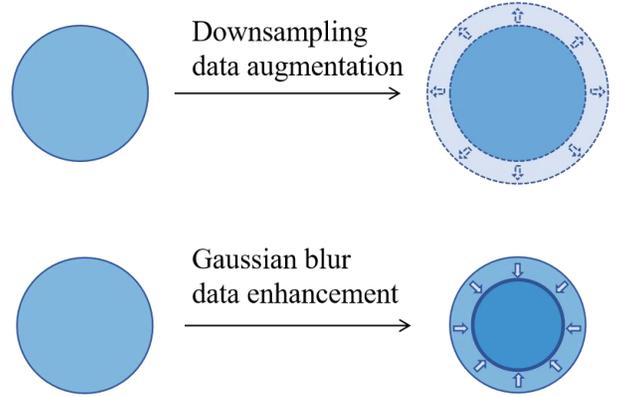

Figure 2: The model, after being trained with downsampling and Gaussian blur, will show changes similar to those illustrated in the figure when the visualization of predicted data results is performed.

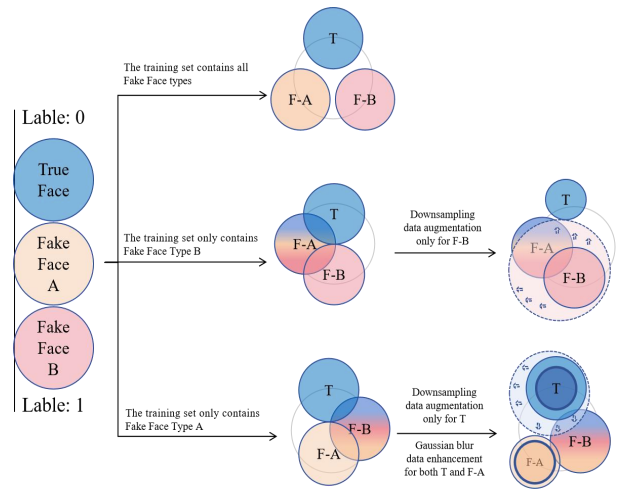

Figure 3: In the Unified Physical-Digital Face Attack Detection competition, we have a total of three tracks, and the data among the three tracks are not shared. For genuine face data, the data is completely consistent across all three tracks. However, for attack face data, Track 1 possesses all types of fake face data, and its test set will also include all types of fake face data. Conversely, Tracks 2 and 3 each contain only one type of fake face data in their training sets, but they are required to validate another type of fake face data that is not included in the training set on their test sets.

Additionally, since the visualization of predicted data through changes in radius only provides an intuitive feeling, we still need a means to perform quantitative analysis with specific data. It is not the case that a larger radius in the data domain necessarily indicates more dispersed data. Therefore, we have defined a standard value *Normalize* Eq. (6) and a density value *Density* Eq. (7) to indicate the concentration of data within the data domain.

$$Normalize = \frac{Pred_{center} - Min(Preds)}{Max(Preds) - Pred_{center}} \quad (6)$$

$$Density = \frac{\sum_{n=1}^{N} Abs(Preds_n - Pred_{center}) / Radius}{N} \quad (7)$$

Ideally, $Normalize$ should be close to 1. If it is very large or very small, it indicates the presence of extremely difficult samples in the data domain. And $Density$, the smaller it is, the more concentrated the distribution; conversely, the larger it is, the more dispersed the distribution.

Therefore, if downsampling are performed, the change of $Normalize$ will be more obvious. And if Gaussian blur is applied, the $Density$ is significantly reduced.

### 3.3. Study on threshold setting

After the model training is complete, another challenge is how to find an appropriate threshold on the validation set in hopes that the model will perform well in actual deployment. The training set is prone to overfitting and may result in extreme classification scenarios. Even if the prediction center points of the validation set and training set are consistent, the sizes of the data domains they contain may differ. Using traditional methods such as ACER to determine the threshold may not necessarily represent the optimal choice for FAS tasks.

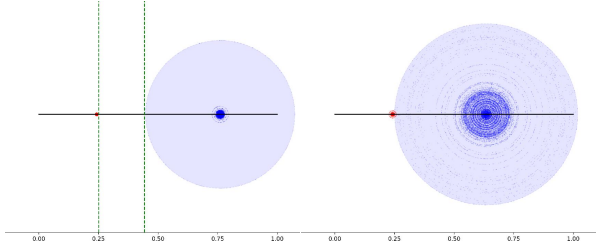

Figure 4: The figure shows a set of visualized images of predicted scores for the training and test sets. The left side represents the training set, while the right side is the test set. In the left image, the green vertical lines represent the boundaries of the two data domains. It is not difficult to see that due to the high distinction between positive and negative samples in the training set, any score within the region from the left vertical line to the right vertical line can serve as a threshold for ACER in the training set. However, by comparing with the test set, it is evident that not every threshold will achieve the expected results on the test set.

In Figure 4, we should be able to employ more sophisticated means to select thresholds. Referring to the method of choosing central points in Section 3.1, we can see that the selection method for $Threshold_{ACER}$ Eq. (8) corresponds to calculating the median value, representing an expectation that the proportions of genuine and fake samples correctly predicted are equal.

$$\frac{\sum_{n=1}^{N} Bool(PredsT_n < Threshold_{ACER})}{N} = \frac{\sum_{m=1}^{M} Bool(PredsF_n > Threshold_{ACER})}{M} \quad (8)$$

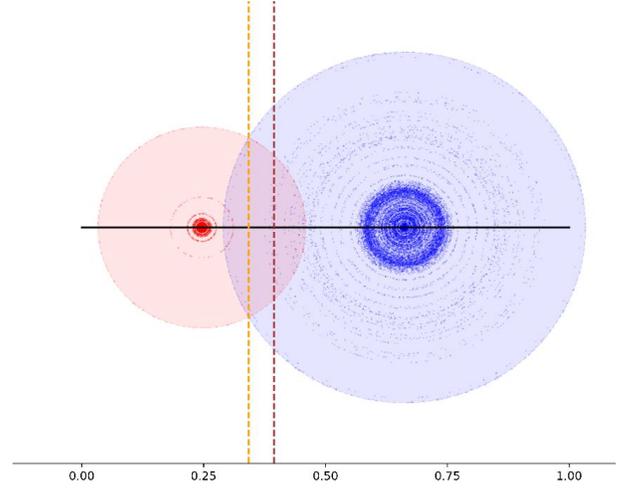

Figure 5: The figure illustrates another common situation encountered during training, where the training set cannot fully converge, and different data domains consistently have overlapping regions.

In Figure 5, when two predicted data domains intersect, the threshold corresponding to the point of intersection of the domains can be calculated using the cosine theorem. If the two data zones do not intersect, it is defined as the intermediate value between the two data and the boundary.

Alternatively, a balanced threshold $Threshold_{balance}$ Eq. (9) could be employed using the point of balance between the distances reached by the two data domains.

$$Threshold_{balance} \in [PredN_{max}, PredT_{min}]$$

$$\sum_{n=1}^{N} Threshold_{balance} - PredsF_n = \sum_{m=0}^{M} PredsT_m - Threshold_{balance} \quad (9)$$

Although $Threshold_{ACER}$ has proven its effectiveness through long-term practice, we can still introduce more threshold selection methods to ensure the model exhibits its due performance in actual deployment.

### 4. Experiments

In this section, we conducted extensive experiments based on data from the Unified Physical-Digital Face Attack Detection competition and the Snapshot Spectral Imaging Face Anti-spoofing contest. Next, in Sec 4.1, we will introduce the data distribution and testing criteria for the Unified Physical-Digital Face Attack Detection competition. In Sec 4.2, we will present our model performance without data augmentation and after downsampling and Gaussian blur, followed by data analysis. In Sec 4.3, we will show the data distribution and testing criteria for the Snapshot Spectral Imaging Face Anti-spoofing contest. Sec 4.4 will provide visualized graphs of prediction scores after training with different data processing methods and demonstrate the specific performance of using different thresholds on the model.

### 4.1. Datasets and Metrics for competitions 1

**Unified physical-digital Attack dataset**, namely UniAttackData. The dataset consists of 1,800 participations of 2 and 12 physical and digital attacks, respectively, resulting in a total of 29, 706 videos.Protocol 1aims to evaluate under the unified attack detection task. Unlike the classical single-class attack detection, the unified attack data protocol contains both physical and digital attacks. The huge intra-class distance and diverse attacks bring more challenges to the algorithm design. Protocol 2 evaluates the generalization to "unseen" attack types. The large differences and unpredictability between physical-digital attacks pose a challenge to the portability of the algorithms.

they use the "leave-one-type-out testing" approach to divide Protocol 2 into twosub-protocols, where the test set for each self-sub-protocol is an unseen attack type. the test set of protocol 2.1 contains only physical attacks that have not been seen in the training and development sets, and the test set of protocol 2.2 contains only digital attacks that have not been seen in the training and development set.

**Performance Metrics**. Three metrics, i.e., Attack Presentation Classification Error Rate (APCER), Bona Fide Presentation Classification Error Rate (BPCER), and Average Classification Error Rate (ACER) [14] are utilized for performance comparison. They can be formulated as

$$APCER = \frac{FP}{TN + FP}$$

$$BPCER = \frac{FN}{FN + TP}$$

$$ACER = \frac{APCER + BPCER}{2} \qquad (10)$$

where $FP, FN, TN$ and $TP$ denote the false positive, false negative, true negative and true positive sample numbers, respectively. ACER is used to determine the final ranking in ChaLearn Face Anti-spoofing Attack Detection Challenge@CVPR2024.

### 4.2. Implementation Details for competitions 1

One of the purposes of this paper is to tackle cross-domain issues by validating data augmentation methods; therefore, we did not engage in further selection of backbone networks. In the end, we solely utilized *ConvNeXtv2-B*[19] as our backbone network. Given the complexity of cross-domain tasks, techniques such as *label smoothing*[20] and *cut-mix*[21] were also incorporated into the training but were not used as variables for comparison. The training parameters across all protocols are nearly consistent.

Considering that Protocol 1 (*p1*) includes all types of attack data, we did not perform any additional data augmentation on *p1*. However, *p2* involves cross-domain attack data, and training with the original data would lead to cross-domain data being almost indistinguishable. As shown in Figure 3, we applied different data augmentation techniques to *p2.1* and *p2.2* respectively. We believe there exists a relative distance relationship between live-advs/digi-phys. Therefore, by applying data augmentation to a certain category of data to change its data distribution domain, it is possible to distinguish cross-domain data. Specifically, for *p2.1* data, we performed 2x downsampling data augmentation on live data and then applied Gaussian blur to all data during training. For *p2.2* data, we conducted 2x/4x/8x downsampling data augmentation on phys attack data. The reason we do not apply uniform data augmentation to all types of data is twofold. On one hand, the downsampling operation, while expanding the data distribution domain, also introduces noise, and we prefer not to introduce excessive irrelevant information. On the other hand, due to the application of the *cut-mix* technique, augmenting one category of data can influence the training outcomes of other categories. Table 1 shows the changes in the predicted scores of models on their respective datasets before and after data augmentation. Table 2 presents the changes in the distribution of predicted scores across different datasets through model cross-inference.

Urthermore, through the $Radius$, $Normalize$ and $Density$, Tables 3 and 4 also intuitively show the numerical changes corresponding to different operations. Clearly, the downsampling operation makes the data domains more expansive, while Gaussian blur tends to concentrate the data within the domains.

It can be observed that the resize data augmentation operation, after expanding the data domain, can significantly alter the balance point between classes. However, it does not make a noticeable change to the density of the data domain itself. On the other hand, the Gaussian blur operation can significantly change the density of the data itself, making it more cohesive.

Considering that each task has a different ratio of sample quantities, this may be one of the reasons why the resize operation has varying degrees of impact on the balance point.

|  | Model-P1 | Model-P2.1 (Orig) | Model-P2.1 (Extend) | Model-P2.2 (Orig) | Model-P2.2 (Extend) |
|---|---|---|---|---|---|
| **Correct Dataset Train** | | | | | |
| **Correct Dataset Dev** | | | | | |

Table 1: The visualization results of the predicted scores on their corresponding training and validation sets for models in each track.

|  | Model-P1 | Model-P2.1 (Orig) | Model-P2.1 (Extend) | Model-P2.2 (Orig) | Model-P2.2 (Extend) |
|---|---|---|---|---|---|
| **Dataset Train-P1** | | | | | |
| **Dataset Train-P2.1** | | | | | |
| **Dataset Train-P2.2** | | | | | |
| **Dataset Dev-P1** | | | | | |
| **Dataset Dev-P2.1** | | | | | |
| **Dataset Dev-P2.2** | | | | | |

Table 2: The visualization results of the predicted scores for models in each track on different training and validation sets.

| Dataset train | Model P2.1 | Radius | Normalize | Density |
|---|---|---|---|---|
| Live Face | Orig | 0.16387 | 0.06212 | 0.00492 |
| | Extend | **0.17958** | **0.03163** | **0.00293** |
| All Fake Face | Orig | 0.17167 | 4.58766 | 0.32506 |
| | Extend | 0.17984 | 6.51122 | 0.22755 |
| Phys Fake Face | Orig | 0.13114 | 1.63312 | 0.63237 |
| | Extend | 0.15308 | 2.63840 | **0.43475** |
| Adv/Digital Fake Face | Orig | 0.20740 | 16.86177 | 0.00214 |
| | Extend | 0.20267 | 16.50409 | **0.00181** |

Table 3: The changes in predicted scores for each dataset before and after the data augmentation operation on p2.1 data.

| Dataset train | Model P2.2 | Radius | Normalize | Density |
|---|---|---|---|---|
| Live Face | Orig | 0.02316 | 2.76702 | 0.06564 |
| | Extend | 0.22304 | 0.12164 | 0.05546 |
| All Fake Face | Orig | 0.15432 | 0.70159 | 0.52670 |
| | Extend | 0.22453 | 6.46501 | 0.04794 |
| Phys Fake Face | Orig | 0.14907 | 3.39799 | 0.01001 |
| | Extend | **0.23244** | **16.75847** | 0.01381 |
| Adv/Digital Fake Face | Orig | 0.18679 | 0.18797 | 0.12584 |
| | Extend | **0.19596** | **3.35089** | 0.16948 |

Table 4: The changes in predicted scores for each dataset before and after the data augmentation operation on p2.1 data.

### 4.3. Datasets and Metrics for competitions 2

**HySpeFAS**, which contains 6760 hyperspectral images reconstructed from SSI images by TwIST algorithm. Each hyperspectral image contains 30 spectral channels. Meanwhile, as illustrated in Figure 6, the organizers have provided a proprietary algorithm for linearly combining the 30-channel data into 3-channel PNG format images intended for visualization. Participants are allowed to train and perform inference using these three-channel PNG images instead of the entire 30 channels of data. This competition aims to encourage the exploration of spectral anti-spoofing algorithms suitable for SSI images, and to promote research on new spectroscopic sensor face anti-spoofing algorithms.

**Performance Metrics**, There is no fixed algorithm for confirming the threshold. Contestants must perform a binary classification of the predicted data based on the threshold they select. Finally, the error rate is determined using the method for calculating ACER.

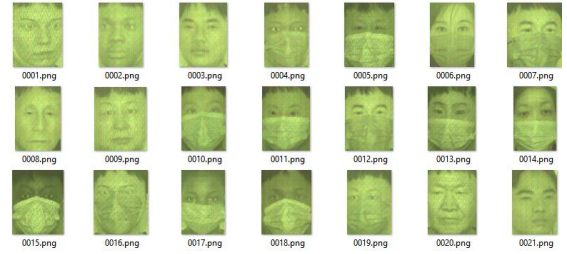

Figure 6: The PNG images are a condensed version of the MAT data, created by applying a linear transformation to reduce the image channels from 30 down to 3. This conversion is designed solely to facilitate easier visualization.

### 4.4. Implementation Details for competitions 2

Similar to the training configuration for competition 1 in 4.2, we utilized the *ConvNeXtv2-B* network with a pretrained model as our training backbone too, and we incorporated training enhancement techniques such as *label smoothing* and *cut-mix*. Considering that the PNG data is a condensed version of the original data, excessive preprocessing could potentially distort the original data distribution. Therefore, during data augmentation, we only subjected the negative samples to 2x downsampling with a 50% probability. In the model training phase, we solely relied on simple color jittering and horizontal flipping for data enhancement, along with applying Gaussian blur to all the data.

According to Tables 5 and 6, this also shows the impact of downsampling and Gaussian blur on the data domain. Based on the visualization results and the statistical data in Table 7 and Figure 7, it is not difficult to see that the ACER value corresponding to the training set may not achieve the same reliable effect on the validation set. Therefore, we finally chose to use the balanced threshold as the model's threshold selection. The effect of the balanced threshold may not necessarily be better than the threshold corresponding to ACER, but it can help us choose a more reasonable threshold for the model.

In Table 6, it also can be seen that the density of the data domain subjected to Gaussian blur without downsampling augmentation has not changed significantly. This suggests that the original data distribution is already very compact, with no space for further consolidation. Therefore, it is necessary to perform downsampling in order to enhance generalization.

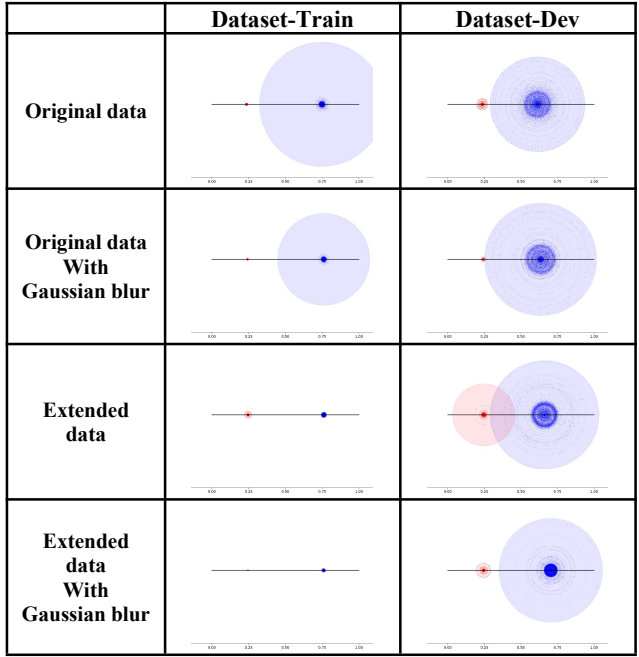

Table 5: Visualization of model prediction scores before and after using different data augmentation methods.

| SSI Data Type | Dataset-Train | Radius | Normalize | Density |
|---|---|---|---|---|
| Live Face | Original data | 0.42396 | 21.29378 | 0.01761 |
| | Original data With Gaussian blur | 0.31287 | 12.10329 | 0.0175 |
| | Extended data | 0.01877 | 0.8796 | 0.23657 |
| | Extended data With Gaussian blur | 0.01315 | 0.67605 | 0.21732 |
| Fake Face | Original data | 0.00878 | 0.98633 | 0.17133 |
| | Original data With Gaussian blur | 0.00639 | 1.0597 | 0.17594 |
| | Extended data | 0.02269 | 0.43808 | 0.05105 |
| | Extended data With Gaussian blur | 0.00324 | 0.99691 | 0.1781 |

Table 6: R/N/D corresponding to each enhancement method of HySpeFAS data.

| | Threshold | Dataset-Train | | | Dataset-Dev | | |
|---|---|---|---|---|---|---|---|
| | | FPR(%) | TPR(%) | ACER(%) | FPR(%) | TPR(%) | ACER(%) |
| Fake border | 0.24815 | 0.0 | 100.0 | 0.0 | 3.434 | 100.0 | 1.717 |
| Live border | 0.74673 | 0.0 | 100.0 | 0.0 | 0.0 | 7.212 | 3.606 |
| Cross point | 0.49744 | 0.0 | 100.0 | 0.0 | 0.0 | 98.558 | 0.725 |
| Balance point | 0.31795 | 0.0 | 100.0 | 0.0 | 0.0 | 100.0 | 0.0 |
| Acer-left point | 0.24815 | 0.0 | 100.0 | 0.0 | 3.434 | 100.0 | 1.717 |
| Acer-right point | 0.74673 | 0.0 | 100.0 | 0.0 | 0.0 | 7.212 | 3.606 |

Table 7: The thresholds selected in different ways on the train dataset, and the corresponding FPR/TPR/ACER of these thresholds on the train dataset and dev dataset.

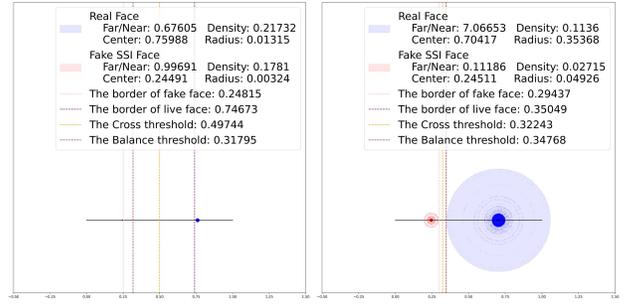

Figure 7: The left graph shows different threshold selection methods and their corresponding values on the train dataset, while the right graph shows different threshold selection methods and their corresponding values on the dev dataset. It can be seen that the results using the $Threshold_{balance}$ are the closest.

## 5. Conclusion

In this paper, we conducted a detailed analysis of the specific effects of downsampling data augmentation and using Gaussian blur for data enhancement through a visualization method. It shows that good results can be achieved in both cross-domain and within-domain tasks. Additionally, we proposed a new method for threshold setting. Applying these methods, we achieved second place in both the Unified Physical-Digital Face Attack Detection competition and the Snapshot Spectral Imaging Face Anti-spoofing contest of ChaLearn Face Anti-spoofing Attack Detection Challenge@CVPR2024.